\documentclass[runningheads]{llncs}
\usepackage{graphicx}
\usepackage{tikz}
\usetikzlibrary{decorations.pathreplacing}
\usepackage{comment}
\usepackage{amsmath,amssymb} 
\usepackage{color}

\usepackage[utf8]{inputenc} 
\usepackage[T1]{fontenc}    
\usepackage{hyperref}       
\usepackage{url}            
\usepackage{booktabs}       
\usepackage{amsfonts}       
\usepackage{nicefrac}       
\usepackage{microtype}      
\usepackage{xcolor}
\usepackage{array}
\usepackage{multirow}

\usepackage{pifont}
\usepackage{stfloats}
\usepackage{subcaption}
\usepackage{wrapfig}
\usepackage[draft,footnote,nomargin]{fixme}
\usepackage{orcidlink}
\usepackage{ulem}

\newcommand{\sectionref}[1]{Section~\ref{sec:#1}}
\newcommand{\figref}[1]{Figure~\ref{fig:#1}}
\newcommand{\tableref}[1]{Table~\ref{tab:#1}}
\newcommand{\equationref}[1]{Equation~(\ref{eq:#1})}

\usepackage[accsupp]{axessibility}  

\begin{document}
\pagestyle{headings}
\mainmatter
\title{MOTCOM: The Multi-Object Tracking Dataset Complexity Metric} 

\author{Malte Pedersen\inst{1}\orcidlink{0000-0002-2941-9150} \and
Joakim {Bruslund Haurum}\inst{1}$^,$\inst{2}\orcidlink{0000-0002-0544-0422} \and
Patrick Dendorfer\inst{3}\orcidlink{0000-0002-4623-8749} \and\\
Thomas {B. Moeslund}\inst{1}$^,$\inst{2}\orcidlink{0000-0001-7584-5209}}
\authorrunning{M. Pedersen et al.}
%
\institute{Aalborg University, Denmark\and
Pioneer Center for AI, Denmark\and
Technical University of Munich, Germany}
\maketitle

\begin{abstract}
There exists no comprehensive metric for describing the complexity of Multi-Object Tracking (MOT) sequences.
This lack of metrics decreases explainability, complicates comparison of datasets, and reduces the conversation on tracker performance to a matter of leader board position.
As a remedy, we present the novel MOT dataset complexity metric (MOTCOM), which is a combination of three sub-metrics inspired by key problems in MOT: occlusion, erratic motion, and visual similarity.
The insights of MOTCOM can open nuanced discussions on tracker performance and may lead to a wider acknowledgement of novel contributions developed for either less known datasets or those aimed at solving sub-problems.

We evaluate MOTCOM on the comprehensive MOT17, MOT20, and MOTSynth datasets and show that MOTCOM is far better at describing the complexity of MOT sequences compared to the conventional \textit{density} and \textit{number of tracks}. Project page at \url{https://vap.aau.dk/motcom}.
\end{abstract}

\section{Introduction}\label{intro}
Tracking has been an important research topic for decades with applications ranging from autonomous driving to fish behavior analysis \cite{uhlmann1992algorithms,luo2020multiple,gade2018constrained,perez2014idtracker}.
The aim is to acquire the full spatio-temporal trajectory of an object of interest, but missing or inaccurate detections can make this a complicated task.
When more objects are present in the scene simultaneously it is termed a multi-object tracking (MOT) problem and an additional task is to keep the correct identities of all objects throughout the sequence.

\begin{figure}[t]
    \centering
    \includegraphics[width=\linewidth]{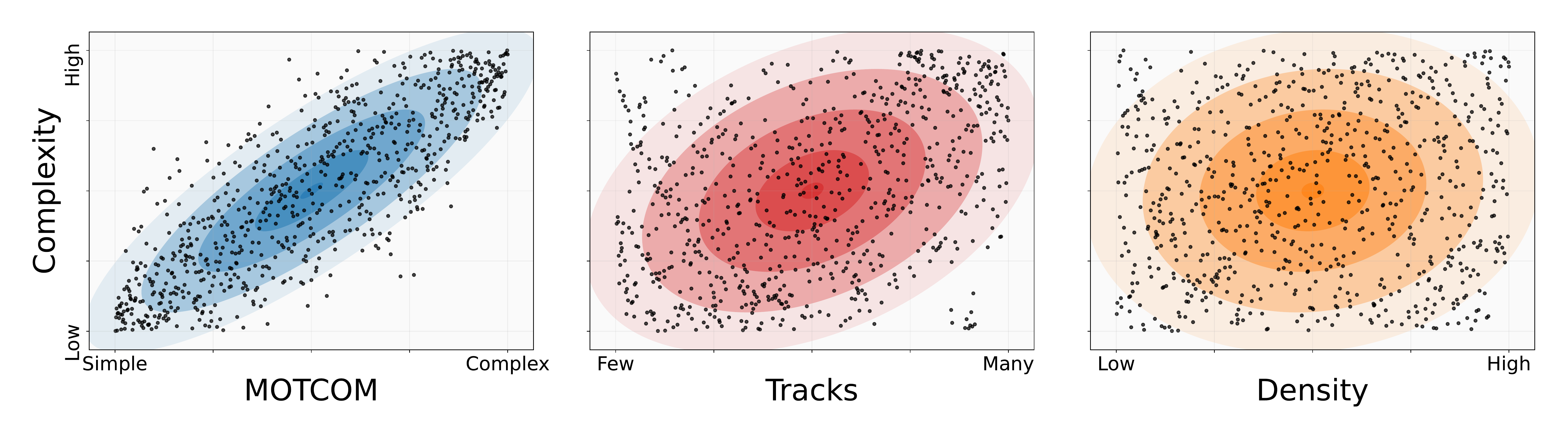}
    \caption{Comparing the capability of the proposed MOTCOM metric against the conventional metrics (\textit{number of tracks} and \textit{density}) for describing MOT sequence complexity. The shared y-axis shows a HOTA \cite{luiten2020hota} rank-based proxy for the ground truth complexity of the MOTSynth sequences \cite{fabbri2021motsynth}. The x-axes show the corresponding rank determined by each of the three metrics. The correlation between the complexity and MOTCOM is clearly stronger compared to both \textit{tracks} and \textit{density}. More details can be found in \sectionref{evaluation}.}
    \label{fig:front}
\end{figure}

During the previous decade there has been an increase in the development of publicly available MOT datasets \cite{geiger2012we,milan2016mot16,dendorfer2020motchallenge,Sun_2020_CVPR,Argoverse}.
However, there has been no attempt to objectively describe the complexity of a dataset or its sequences except for using simple statistics like \textit{density} and \textit{number of tracks}, which are neither adequate nor explanatory, see \figref{front}.
When a new dataset emerges, the community needs objective metrics to be able to characterize and discuss the dataset with respect to existing datasets, otherwise, ‘gut feeling’ and ‘popularity vote’ will rule.
Furthermore, the absence of an objective MOT sequence complexity metric hinders an informed conversation on the capabilities of trackers developed for different datasets.
Nowadays, it is important to rank high on popular MOT benchmark leaderboads in order to gain the attention of the community.
This may hinder the acknowledgement of novel solutions that solve sub-problems of MOT particularly well and underrate solutions developed on less popular datasets.
We expect that a descriptive and explanatory metric can help remedy these issues.

The literature suggests that there are three main factors that make MOT tasks difficult to solve~\cite{Bergmann_2019_ICCV,Pedersen_2020_CVPR,andriyenko2011analytical,andriyenko2011multi,luo2020multiple}; namely, occlusion, erratic motion, and visual similarity.
We hypothesize that the complexity of MOT sequences can be expressed by a combination of the aforementioned three factors for which we need to construct explicit metrics.
Therefore, in this paper we propose the first-ever individual sub-metrics for describing the complexity of the three sub-problems and a unified quantitative MOT dataset complexity metric (MOTCOM) as a combination of these sub-metrics.
In \figref{front}, we illustrate that MOTCOM is far better at estimating the complexity of the sequences of the recent MOTSynth dataset \cite{fabbri2021motsynth} compared to the commonly used \textit{number of tracks} and \textit{density}.

The main contributions of our paper are as follows:
\begin{enumerate}
    \item The novel metric MOTCOM for describing the complexity of MOT sequences.
    \item Three sub-metrics for describing the complexity of MOT sequences with respect to occlusion, erratic motion, and visual similarity.
    \item We show that the conventional metrics \textit{number of tracks} and \textit{density} are not strong indicators for the complexity of MOT sequences.
    \item We evaluate the capability of MOTCOM and demonstrate its superiority against \textit{number of tracks} and \textit{density}.
\end{enumerate}
In the next section, we describe and analyse the three sub-problems followed by a presentation of the proposed metrics.
In the remainder of the paper, we demonstrate and discuss how the metrics can describe and explain the complexity of MOT sequences.

\section{Related Work}
The majority of recent trackers utilize the strong performance of deep learning based detectors, e.g., by following the tracking-by-detection paradigm \cite{xu2019spatial,bewley2016simple,wojke2017simple}, tracking-by-regression \cite{Bergmann_2019_ICCV}, through joint training of the detection and tracking steps \cite{peng2020chained,zhou2020tracking}, or as part of an association step \cite{pang2021quasi,lu2020retinatrack,zhang2021fairmot}. 
Trackers like Tracktor \cite{Bergmann_2019_ICCV}, Chained-Tracker \cite{peng2020chained}, and CenterTrack \cite{zhou2020tracking} rely on spatial proximity which makes them vulnerable to sequences with extreme motion and heavy occlusion.
At the other end of the spectrum are trackers like QDTrack \cite{pang2021quasi}, RetinaTrack \cite{lu2020retinatrack}, and FairMOT \cite{zhang2021fairmot} which use visual cues for tracking.
They are optimized toward tracking visually distinct objects and are not to the same degree limited by erratic motion or vanishing objects but instead sensitive to weak visual features.
This indicates that the design of trackers is centered around three core problems: occlusion, erratic motion, and visual similarity.
Below, we dive into the literature regarding these problems followed by insights on dataset complexity.

\subsubsection{Occlusion.}
Occlusions can be difficult to handle and they are often simply treated as missing data \cite{andriyenko2011multi}.
However, in scenes were the objects have weak or similar visual features this can be harmful for the tracking performance \cite{andriyenko2011analytical,milan2013continuous,stadler2021improving}.

Most authors state that a higher occlusion rate makes tracking harder \cite{Cao2020,liu2019model,luo2014bi}, but they seldom quantify such statements.
An exception is the work proposed by Bergmann et al. \cite{Bergmann_2019_ICCV} where they analyzed the tracking results with respect to object visibility, the size of the objects, and missing detections.
Moreover, Pedersen et al. \cite{Pedersen_2020_CVPR} argued that the number of objects is less critical than the amount and level of occlusion when it comes to multi-object tracking of fish.
They described the complexity of their sequences based on occlusions alone.

\subsubsection{Erratic Motion.}
Prior information can be used to predict the next state of an object which minimizes the search space and hence reduces the impact of noisy or missing detections.
A linear motion model assuming constant velocity is a simple, but effective method for predicting the movement of non-erratic objects like pedestrians \cite{luo2020multiple,milan2013continuous}.
In scenes that include camera motion or complex movement more advanced models may improve tracker performance.
Pellegrini et al. \cite{Pellegrini2009} proposed incorporating human social behavior into their motion model and Kratz et al. \cite{kratz2010tracking} proposed utilizing the movement of a crowd to enhance the tracking of individuals.
A downside of many advanced motion models is an often poor ability to generalize to other types of objects or environments.

\subsubsection{Visual Similarity.}
Visual cues are commonly used in tracklet association and re-identification and are well studied for persons \cite{ye2021personreid}, vehicles \cite{khan2019vehiclereid}, and animals \cite{schneider2019animalreid} such as zebrafish \cite{haurum2020zebrafishreid} and tigers \cite{schnedier2020animalreid}. 
Modern trackers often solve the association step using CNNs, like Siamese networks, based on a visual affinity model \cite{Bergmann_2019_ICCV,leal2016learning,xiang2015learning,yin2020unified}.
Such methods rely on visual dissimilarity between the objects. 
However, tracklet association becomes more difficult when objects are hard to distinguish purely by their appearance.

\subsubsection{Dataset Complexity.}
Determining the complexity of a dataset is a non-trivial task.
One may have a ``feeling'' or intuition about which datasets are harder than others, but this is subjective and can differ depending on who you ask, as well as differ depending on the task at hand.
In order to objectively determine the complexity of a dataset, one has to develop a task-specific framework.
An early attempt at this was the suite of 12 complexity measures (c-measures) by Ho and Basu \cite{HoBasu2002}, based on concepts such as inter-class overlap and linear separability.
However, these c-measures are not suitable for image datasets due to unrealistic assumptions, such as the data being linearly separable.
Therefore, Branchaud-Charron et al. \cite{Branchaud-Charron_2019_CVPR} developed a complexity measure based on spectral clustering, where the inter-class overlap is quantified through the eigenvalues of an approximated adjacency matrix.
This approach was shown to correlate well with the CNN performance on several image datasets.
Similarly, Cui et al. \cite{cui2019measuring} presented a framework for evaluating the fine-grainedness of image datasets, by measuring the average distance from data examples to the class centers.
Both of these approaches rely on embedding the input images into a feature space by using, e.g., a CNN, and determining the dataset complexity without any indication of what makes the dataset difficult.

In contrast, dataset complexity in the MOT field has so far been determined through simple statistics such as the number of tracks and density.
These quantities are currently displayed for every sequence alongside other stats such as resolution and frame rate for the MOTChallenge benchmark datasets \cite{dendorfer2020motchallenge}.
The preliminary works of Bergmann et al. \cite{Bergmann_2019_ICCV} and Pedersen et al. \cite{Pedersen_2020_CVPR} have attempted to further explain what makes a MOT sequence difficult by investigating the effect of occlusions. 
However, there is no clear way of describing the complexity of MOT sequences and the current methods have not been verified.

\section{Challenges in Multi-Object Tracking}\label{sec:challenges}
MOT covers the task of obtaining the spatio-temporal trajectories of multiple objects in a sequence of consecutive frames. 
Depending on the specific task, the objects may be represented as 3D points \cite{Pedersen_2020_CVPR}, pixel-level segmentation masks \cite{Voigtlaender_2019_CVPR}, or bounding boxes \cite{milan2013challenges}.
Despite the different representation forms, the concepts of occlusion, erratic motion, and visual similarity apply to all of them and add to the complexity of the sequences.

\subsubsection{Occlusion.}
\begin{figure}[!b]
    \centering
    \begin{subfigure}[t]{0.4\textwidth}
        \centering
        \includegraphics[width=\linewidth]{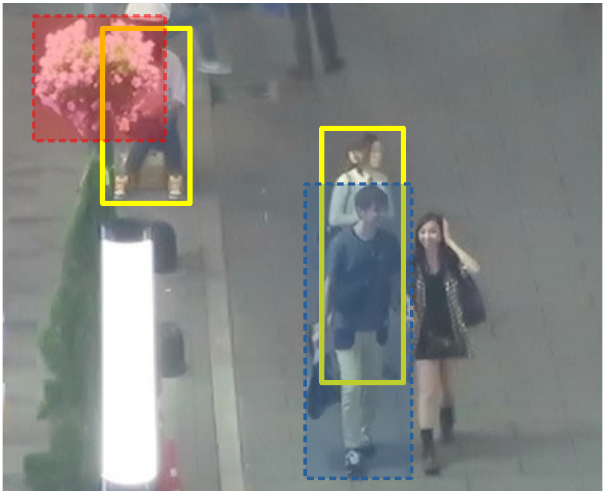}
        \caption{}
        \label{fig:occlusion_types}
    \end{subfigure}
    \qquad \qquad
    \begin{subfigure}[t]{0.45\textwidth}
        \centering
        \includegraphics[width=\linewidth]{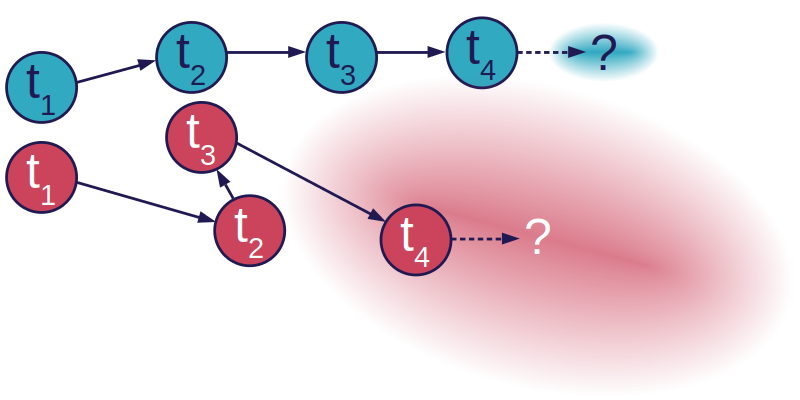}
        \caption{}
        \label{fig:erratic_motion}
    \end{subfigure}
    \caption{a) Sample from MOT17-04 \cite{milan2016mot16}. The yellow boxes illustrate objects partly occluded by scene-occlusion (red) and inter-object-occlusion (blue). b) The blue object displays nearly linear motion, whereas the red object is behaving erratically. The ellipsoids symbolize the confidence of an artificial underlying motion model.}
\end{figure}

Occlusion describes situations where the visual information of an object within the camera view is partially or fully hidden.
There are three types of occlusion: \textit{self-occlusion}, \textit{scene-occlusion}, and \textit{inter-object-occlusion} \cite{andriyenko2011analytical}.
Self-occlusion can reduce the visibility of parts of an object, e.g., if a hand is placed in front of a face, but defining the level of self-occlusion is non-trivial and depends on the type of object.
Scene-occlusion occurs when a static object is located in the line of sight between the camera and the target object, thereby decreasing the visual information of the target.
A scene-occlusion is marked by the red box in \figref{occlusion_types}, where flowers partially occlude a sitting person.

Inter-object-occlusion is typically the most difficult to handle, especially if the objects are of the same type, as the trajectories of multiple objects cross.
An example can be seen in \figref{occlusion_types}, where the blue box marks a person that partially occludes another person. 

\subsubsection{Erratic Motion.}
We use motion as a term for an object's spatial displacement between frames.
This is typically caused by the locomotive behavior of the object itself, camera motion, or a combination.
As the number of factors that influence the observed motion increases, the motion becomes harder to predict.
An example of two objects exhibiting different types of motion is presented in \figref{erratic_motion}.
The blue object moves with approximately the same direction and speed between the time steps.
Predicting the next state of the object seems trivial and the search space is correspondingly small.
On the other hand, the red object behaves erratically and unpredictably while the motion model is less confident as illustrated by the larger search space.
\subsubsection{Visual Similarity.}
The visual appearance of objects can vary widely depending on the type of object and type of scene.
Appearance is especially important when tracking is lost, for example, due to occlusion, and re-identification is a common tool for associating broken tracklets. 
The complexity of this process depends on the visual similarity between objects, but intra-object similarity also plays a role.
As an object moves through a scene, its appearance can change from the perspective of the viewer.
The object may turn around, increase its distance to the camera, or the illumination conditions may change. 
Aside from the visual cues, the object's position is also critical.
Intuitively, it becomes less likely to confuse objects as the spatial distance between them increases.

\section{The MOTCOM Metrics}\label{sec:motcom_metrics}
We propose individual metrics to describe the level of occlusion, erratic motion, and visual similarity for MOT sequences.
Subsequently, we combine these three sub-metrics into a higher-level metric that describes the overall complexity of the sequences.

\subsubsection{Preliminaries}
We define a MOT sequence as a set of frames $F = \{1, 2, \dots\}$ containing a set of objects $K = \{k_1, k_2, \dots\}$. 
The objects do not have to be present in every frame, therefore, we define the set of frames where a given object is present by $F^{k} = \{t_1, t_2, \dots\}$.
The objects present in a given frame $t$ are defined as the set $K^t = \{k | k \in K \land~t \in F^k\}$.
At each frame $t$ an object $k$ is represented by its center-position in image coordinates and the height and width of the surrounding bounding box $k_t = (x,y,h,w)$.
\subsection{Occlusion Metric}
As mentioned in \sectionref{challenges}, occlusion can be divided into three types: self-, scene- and inter-object occlusion.
In order to quantify the occlusion rate in a sequence, one should ideally account for all three types.
However, it is most often non-trivial to determine the level of self-occlusion and it is commonly not taken into account in MOT.
Pedersen et al. \cite{Pedersen_2020_CVPR} used the ratio of intersecting object bounding boxes to determine the inter-object occlusion rate.
Similarly, the MOT16, MOT17, and MOT20 datasets include a visibility score based on the intersection over area (IoA) of both inter- and scene-objects \cite{dendorfer2020motchallenge}, where IoA is formulated as the area of intersection over the area of the target.

Following this trend, we omit self-occlusion and base the occlusion metric, OCOM, on the IoA and compute it as
\begin{equation}
    \mathrm{OCOM} = \frac{1}{|K|} \sum_k^K \bar{\nu}^k,
\end{equation}
where $\bar{\nu}^k$ is the mean level of occlusion of object $k$. $\nu^k_t$ is in the interval $[0,1]$ where 0 is fully visible and 1 is fully occluded.
It is assumed that terrestrial objects move on a ground plane which allows us to interpret their y-values as pseudo-depth and decide on the ordering.
Annotations are needed to calculate the occlusion level for objects moving in 3D.
OCOM is defined in the interval $[0,1]$ where a higher value means more occlusion and a harder problem to solve.

\subsection{Motion Metric}
The proposed motion metric, MCOM, is based on the assumption that objects move linearly when observed at small time steps.
If this assumption is not upheld, it is a sign of erratic motion and thereby a more complex MOT sequence.

Initially, the displacement vector, $P^k_t$, between the object's position in the current and past time step is calculated as
\begin{equation}
    P^k_{t} = p^k_{t} - p^k_{t-\beta},
\end{equation}
where $p_t$ is the position of object $k$ at time $t$, defined by its $x$- and $y$-coordinates, and $\beta$ describes the temporal step size.
When calculating the displacement between two consecutive frames $\beta = 1$.
The displacement vector in the first frame of a trajectory is set to zero and $\beta$ is capped by the first and last frame of a trajectory when the object is not present at time $t\pm\beta$.
\begin{figure}[!t]
    \centering
    \begin{subfigure}[b]{0.25\textwidth}
        \centering
        \includegraphics[width=\textwidth]{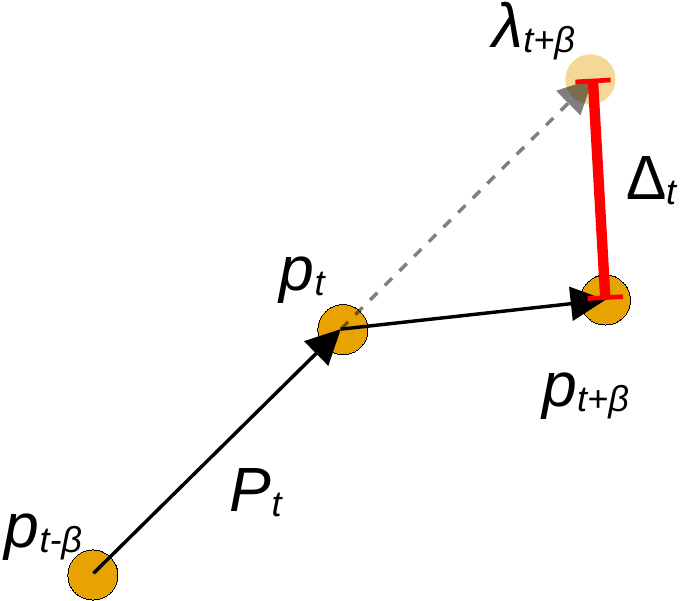}
        \caption{}
        \label{fig:displacement}
    \end{subfigure}
    \qquad\quad
    \begin{subfigure}[b]{0.25\textwidth}
        \centering
        \includegraphics[width=\textwidth]{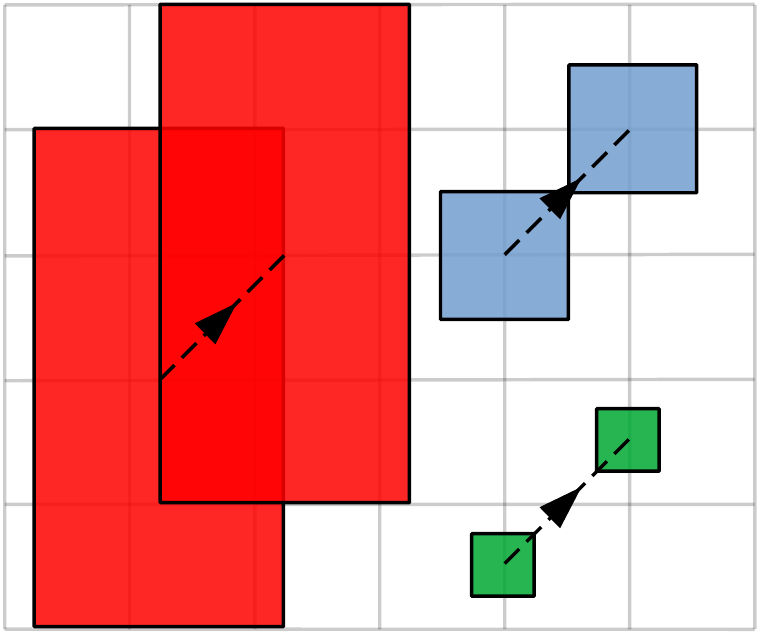}
        \caption{}
        \label{fig:dist_area}
    \end{subfigure}
    \qquad\quad
    \begin{subfigure}[b]{0.25\textwidth}
        \centering
        \includegraphics[width=\textwidth]{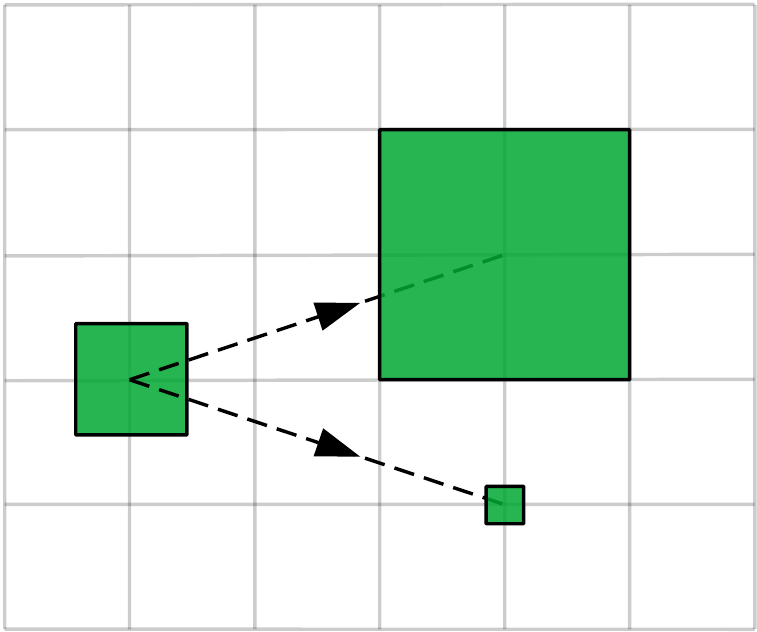}
        \caption{}
        \label{fig:dist_area_change}
    \end{subfigure}
    \caption{a) Illustrative example of how the positional error $\Delta_t$ is calculated as the distance between the true position $p_{t+\beta}$ and estimated position $\lambda_{t+\beta}$. b) The three objects have traveled an equal distance. Relative to their size, the two smaller objects are displaced by a larger amount and the bounding box overlap disappears. c) If the size of an object increases between two time steps the displacement is relatively less important, compared to when the size of the object decreases.}
    \label{fig:displacement_and_area}
\end{figure}

The position in the next time step is predicted using a linear motion model with constant velocity based on the current position and the calculated displacement vector.
The position is predicted by

\begin{equation}
    \lambda^k_{t+\beta} = P^k_{t} + p^k_t.
\end{equation}
The error between the predicted and true position of the object is calculated by
\begin{equation}
    \Delta^k_{t} = \ell_2(p^k_{t+\beta}, \lambda^k_{t+\beta})
\end{equation}
where $\ell_2$ is the Euclidean distance function and a larger $\Delta^k_t$ indicates a more complex motion.
See \figref{displacement} for an illustration of how the displacement error is calculated.
This approach may seem overly simplified, but it encapsulates changes in both direction and velocity.
Furthermore, it is deliberately sensitive to low frame rates and camera motion, as both factors add to the complexity of tracking sequences.

Inspired by the analysis of decreasing tracking performance with respect to smaller object sizes by Bergmann et al. \cite{Bergmann_2019_ICCV}, the size is also taken into consideration.
The combination of size and movement affects the difficulty of predicting the next state of the object.
In \figref{dist_area}, the rectangles are equally displaced but do not experience the same displacement relative to their size. 
Intuitively, if a set of objects are moving at similar speeds, it is harder to track the smaller objects due to their lower spatio-temporal overlap.

\begin{wrapfigure}{r}{0.45\textwidth}
\centering
    \includegraphics[width=\linewidth]{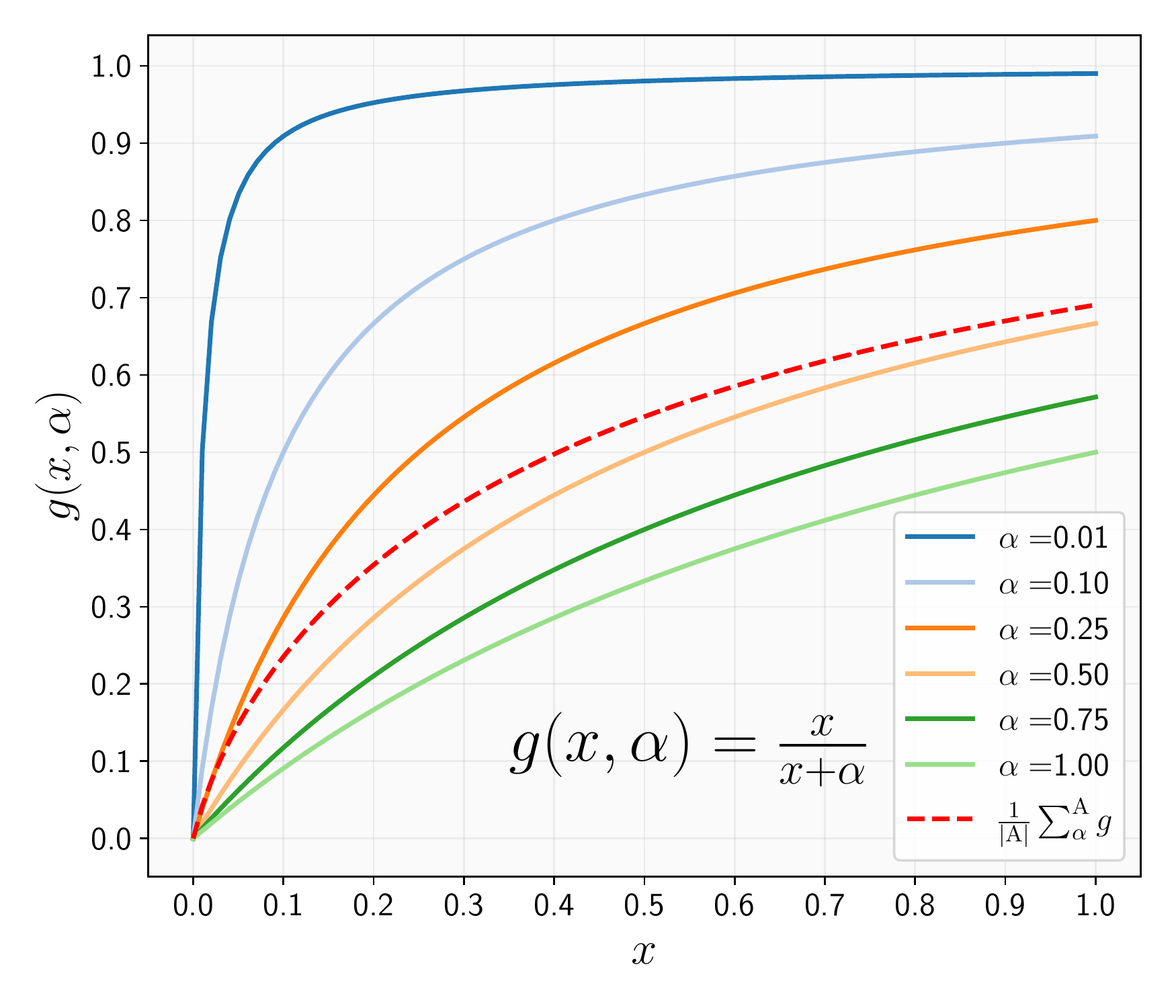}
    \caption{$\alpha$ controls the growth of the function $g(x,\alpha)$ and decides when an output value of 0.5 is reached. The dashed line illustrates $g(x,\alpha)$ when using the average of a set of $\alpha$ values.}
    \label{fig:weighted_func}
\end{wrapfigure}

Accordingly, the motion-based complexity measure is based on the displacement relative to the size of the object.
As illustrated in \figref{dist_area_change}, the size of the object may change between two time steps.
The direction of the change is critical as the displacement is less distinct if the size of the object is increasing, compared to the opposite situation.
Therefore, we multiply the current size of the object with the change in object size to get the transformed object size
\begin{equation}
    \rho^k_t = s^k_t \cdot \frac{s^k_{t+\beta}}{s^k_t}=s^k_{t+\beta},
\end{equation}
where $s_t^k = \sqrt{w^k_{t} \cdot h^k_{t}}$ and $h^k_t$ and $w^k_t$ are the height and width of object $k$ at time step $t$, respectively. 
The motion complexity measure is then calculated as the mean size-compensated displacement across all frames, $F$, and all objects at each frame, $K^t$, and weighted by the log-sigmoid function $g(x,\alpha)$
\begin{equation}    
\mathrm{MCOM} = \frac{1}{|A|} \sum_{\alpha}^{A} g\left( \frac{1}{\sum^K_k|F^k|}\sum_k^K\sum^{F^k}_{t} \frac{\Delta^{k}_{t}}{\rho^k_t} , \alpha \right ),
\end{equation}
where the average of $A = \{0.01,0.02,...,1.0\}$ is used to avoid manually deciding on a specific value for $\alpha$.
The use of the function $g(x,\alpha)$ is motivated by the aim of having an output in the range $[0,1]$, where a higher number describes a more complex motion. 
The function $g(x,\alpha)$ is given by
\begin{equation}\label{eq:weighted_func}
    g(x,\alpha) = \frac{1}{1+e^{-\log(x)}\alpha} = \frac{1}{1+\frac{\alpha}{x}} = \frac{x}{x+\alpha},
\end{equation}
where $\alpha$ affects the gradient of the monotonically increasing function and indicates the point where the output of the function will reach 0.5 as illustrated in \figref{weighted_func}.
The function is designed such that displacements in the lower ranges are weighted higher.
The argument for this choice is based on the assumption that minor increments to an extraordinarily erratic locomotive behavior have less impact on the complexity.
\subsection{Visual Similarity Metric}
In order to define a metric that links an object's visual appearance with tracking complexity, we investigate how similar an object in one frame is compared to itself and other objects in the next frame.
Two objects may look similar, but they cannot occupy the same spatial position.
Therefore, we propose a spatial-aware visual similarity metric called VCOM.

\begin{figure}[!b]
    \centering
    \begin{subfigure}[b]{0.4\textwidth}
        \centering
        \includegraphics[width=\textwidth]{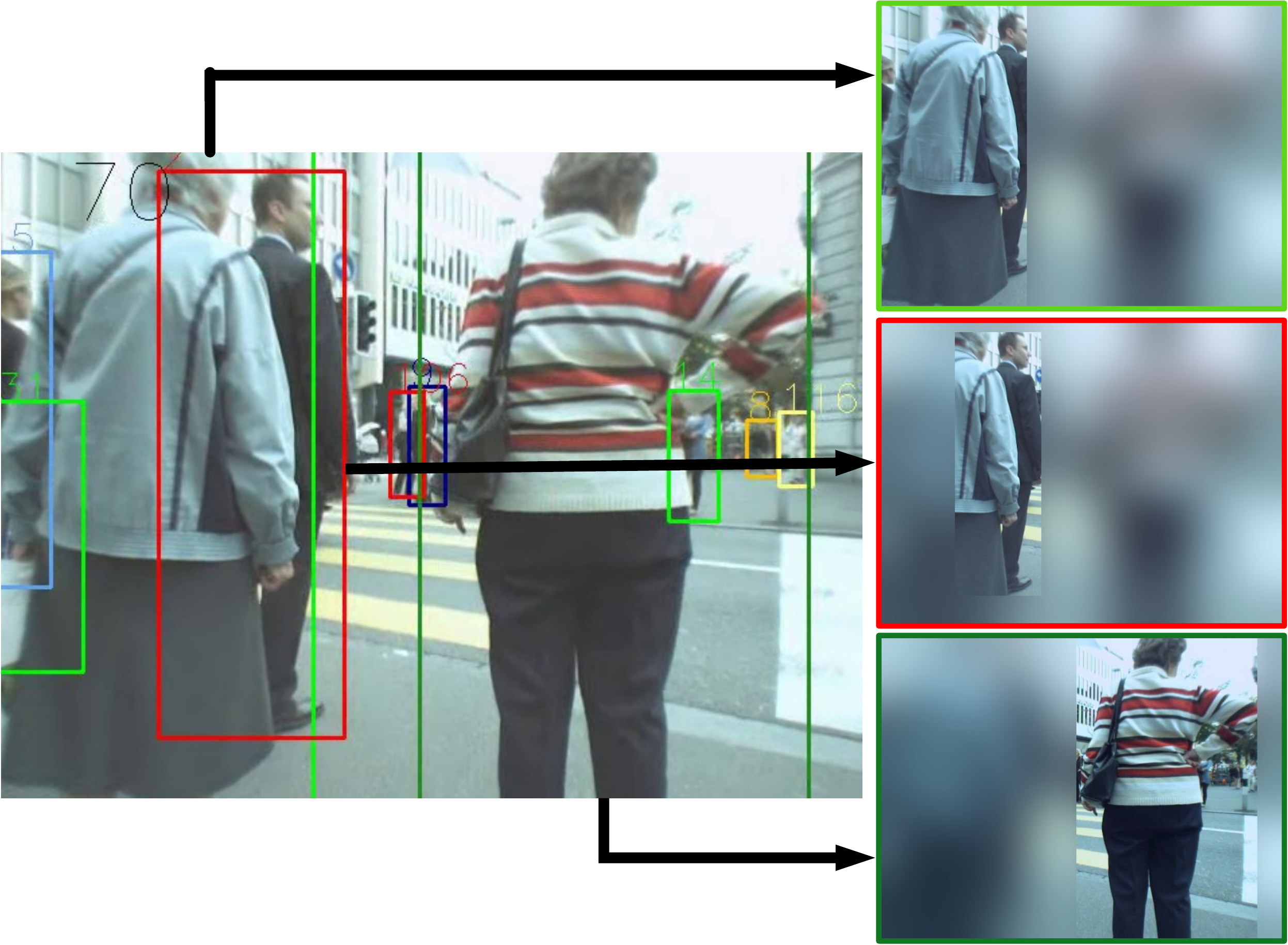}
        \caption{}
        \label{fig:blurred}
    \end{subfigure}
    \qquad \quad
    \begin{subfigure}[b]{0.35\textwidth}
        \centering
        \includegraphics[width=\textwidth]{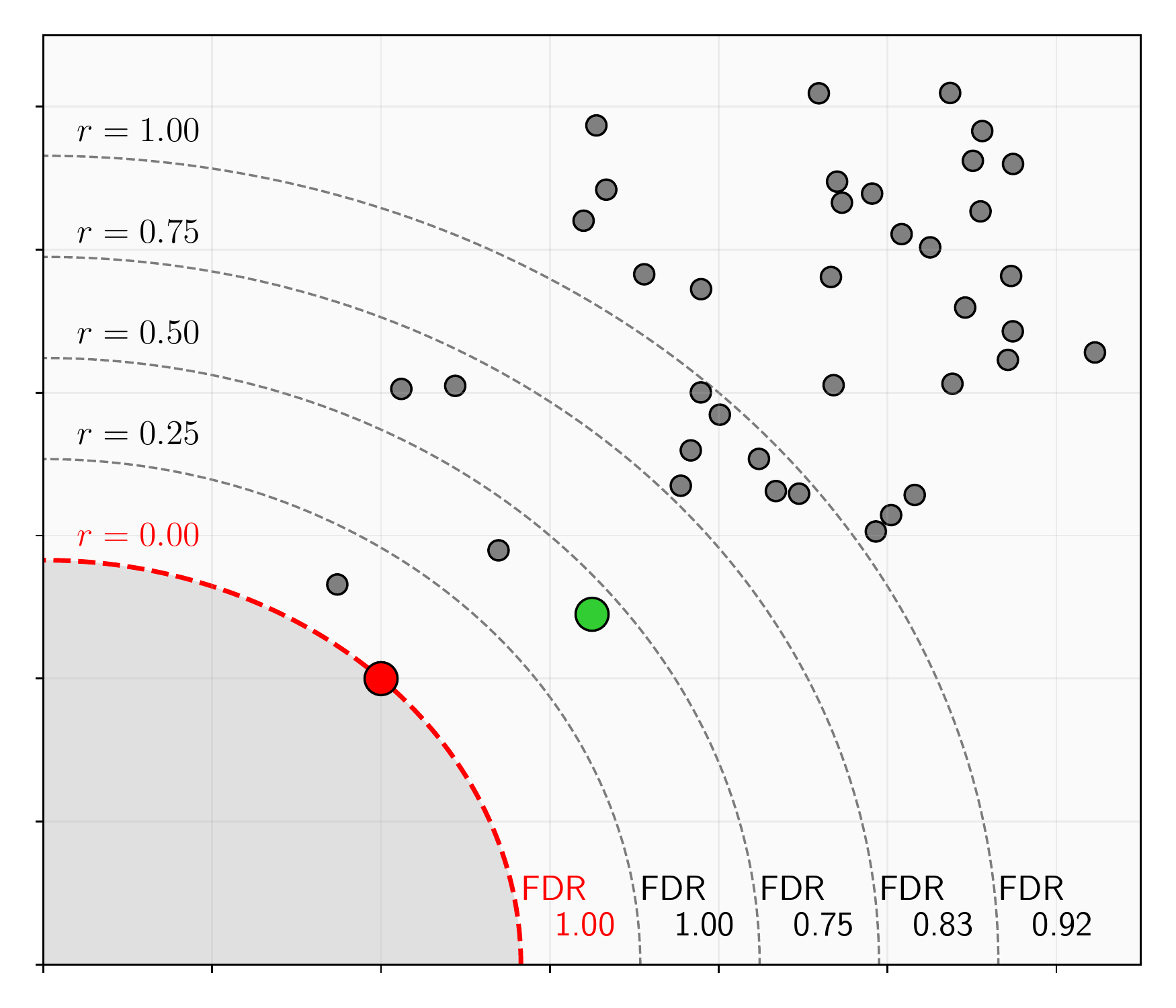}
        \caption{}
        \label{fig:distance}
    \end{subfigure}
    \caption{a) Example showing three images with the object in focus and a blurred background produced from a frame from the MOT17-05 sequence. b) The distance ratio, $r$, affects the FDR when other objects are in the proximity of the target. The red dot is the nearest neighbor, the green dot is the true positive match, and the remaining dots are other objects.}
\end{figure}

VCOM consists of a preprocessing, feature extraction, and distance evaluation step.
For every object $k \in K$ in every frame $t \in F$ an image $I^k_t$ is produced with the object's bounding box in focus and a heavy blurred background.
We blur the image using a discrete Gaussian function, except in the region of the object's bounding box as visualized in \figref{blurred}.

A feature embedding is then extracted from each of the preprocessed images.
As opposed to looking at the bounding box alone, using the entire image allows us to retain and embed spatial information in the feature vector.
The object's location is especially valuable in scenes with similarly looking objects and the blurred background contributes with low frequency information of the surroundings.

We blur the image with a Gaussian kernel with a fixed size of 201 and a sigma of 38 and extract the image features using an ImageNet \cite{ImageNet_2009_CVPR} pre-trained ResNet-18 \cite{ResNet_2016_CVPR} model.
We measure the similarity between the feature vector of the target object in frame $t$ and the feature vectors of all the objects in frame $t+1$ by computing the Euclidean distance.
The uncertainty increases if more objects are located within the proximity of the target. 
Therefore, we do not only look for the nearest neighbor, but rather the number of objects within a given distance, $d(r)$, from the target feature vector
\begin{equation}\label{eq:region}
    d(r) = d_{\mathrm{NN}} + d_{\mathrm{NN}} \cdot r
\end{equation}
where $d_{\mathrm{NN}}$ is the distance to the nearest neighbor and $r$ is a distance ratio.
The ratio is multiplied by the distance to the nearest neighbor in order to account for the variance in scale, e.g., as induced by object resolution or distinctiveness.

An object within the distance boundary that shares the same identity as the target object is considered a true positive (TP) and all other objects are considered false positives (FP).
By measuring the complexity based on the false discovery rate, $\mathrm{FDR} = \frac{\mathrm{FP}}{\mathrm{FP} + \mathrm{TP}}$, we get an output in the range $[0,1]$ where a higher number indicates a more complex task.
An illustrative example of how the FDR is determined based on the distance ratio $r$ can be seen in \figref{distance}.
It is ambiguous to choose a single optimal distance ratio $r$.
Therefore, we calculate VCOM based on the average of distance ratios from the set $R = \{0.01,0.02,...,1.0\}$
\begin{equation}
\mathrm{VCOM} = \frac{1}{|R|}\sum^{R}_{r}\frac{1}{|F|}  \sum_t^F \frac{1}{|K^t|} \sum_k^{K^t} FDR_{d(r)}(k) 
\end{equation}
\subsection{MOTCOM}
Occlusion alone does not necessarily indicate an overwhelming problem if the object follows a known motion model or if it is visually distinct.
The same is true for erratic motion and visual similarity when viewed in isolation.
However, the combination of occlusion, erratic motion, and visual similarity becomes increasingly difficult to handle.

Therefore, we combine the occlusion, erratic motion, and visual similarity metrics into a single MOTCOM metric that describes the overall complexity of a sequence.
MOTCOM is computed as the weighted arithmetic mean of the three sub-metrics and is given by
\begin{equation}
\mathrm{MOTCOM} = \frac{w_{\mathrm{OCOM}}\cdot\mathrm{OCOM}+w_{\mathrm{MCOM}}\cdot\mathrm{MCOM}+w_{\mathrm{VCOM}}\cdot\mathrm{VCOM}}{w_{\mathrm{OCOM}}+w_{\mathrm{MCOM}}+w_{\mathrm{VCOM}}}
\end{equation}
where $w_{\mathrm{OCOM}}$, $w_{\mathrm{MCOM}}$, and $w_{\mathrm{VCOM}}$ are the weights for the three sub-metrics.
Equal weighting can be obtained by setting $w_{\mathrm{OCOM}}=w_{\mathrm{MCOM}}=w_{\mathrm{VCOM}}$, while custom weights may be suitable for specific applications.
During evaluation we weight the sub-metrics equally as we deem each of the sub-problems equally difficult to handle. 

\section{Evaluation}\label{sec:evaluation}
In the following experimental section, we demonstrate that MOTCOM is able to describe the complexity of MOT sequences and is superior to \textit{density} and \textit{number of tracks}.
In order to do this, we compare the estimated complexity levels with ground truth representations.
Such ground truths are not readily available, but a strong proxy can be obtained by ranking the sequences based on the performance of state-of-the-art trackers \cite{leal2017tracking}. 
There exist many performance metrics with two of the most popular being MOTA \cite{stiefelhagen2006clear} and IDF1 \cite{Ristani2016}.
However, we apply the recent HOTA metric \cite{luiten2020hota}, which was proposed in response to the imbalance between detection, association, and localization within traditional metrics.
Additionally, HOTA is the tracker performance metric that correlates the strongest with MOT complexity based on human assessment \cite{luiten2020hota}.
In the remainder of this section, we present the datasets and evaluation metrics we use to experimentally verify the applicability of MOTCOM.

\begin{figure}[b]
\centering
\begin{subfigure}[t]{0.48\textwidth}
\centering
    \includegraphics[width=0.31\linewidth]{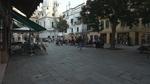}
    \includegraphics[width=0.31\linewidth]{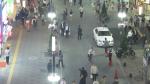}
    \includegraphics[width=0.31\linewidth]{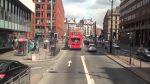}
    \caption{}
\end{subfigure}
\begin{subfigure}[t]{0.48\textwidth}
\centering
    \includegraphics[width=0.31\linewidth]{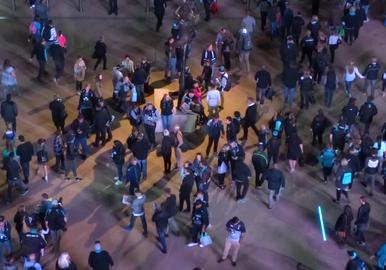}
    \includegraphics[width=0.31\linewidth]{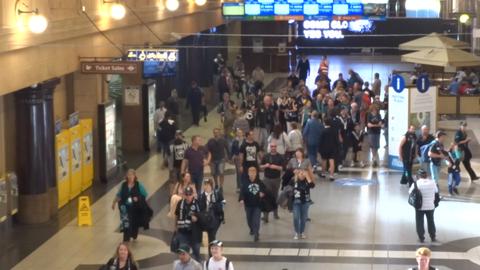}
    \includegraphics[width=0.31\linewidth]{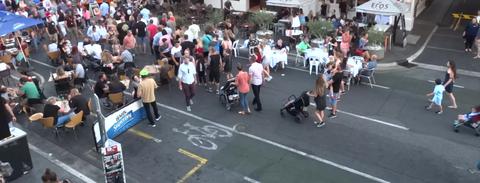}
    \caption{}
\end{subfigure}
\caption{Sample images from a) MOT17  \cite{milan2016mot16} and b) MOT20 \cite{dendorfer2020motchallenge}. MOT17 contains varied urban scenes with and without camera motion. MOT20 contains crowded scenes captured from an elevated point of view and without camera motion.}
\label{fig:motchl_samples}
\end{figure}

\begin{figure}[!b]
\centering
\begin{subfigure}[t]{0.19\textwidth}
\centering
    \includegraphics[width=\linewidth]{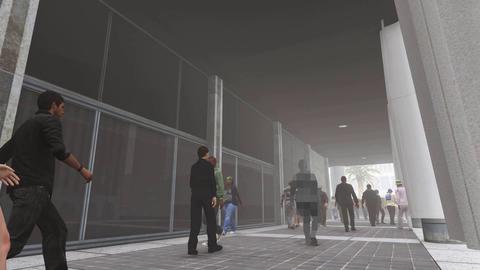}
\end{subfigure}
\begin{subfigure}[t]{0.19\textwidth}
\centering
    \includegraphics[width=\linewidth]{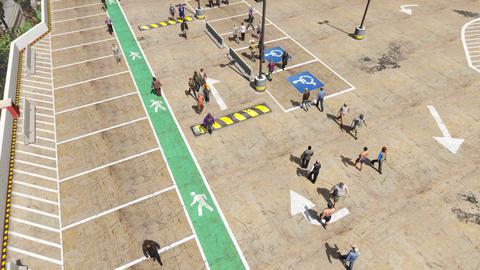}
\end{subfigure}
\begin{subfigure}[t]{0.19\textwidth}
\centering
    \includegraphics[width=\linewidth]{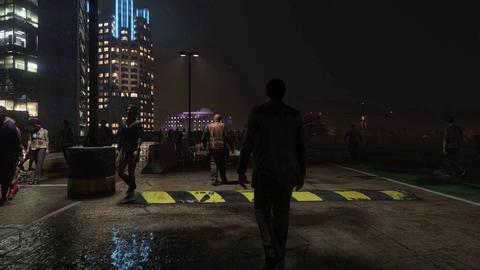}
\end{subfigure}
\begin{subfigure}[t]{0.19\textwidth}
\centering
    \includegraphics[width=\linewidth]{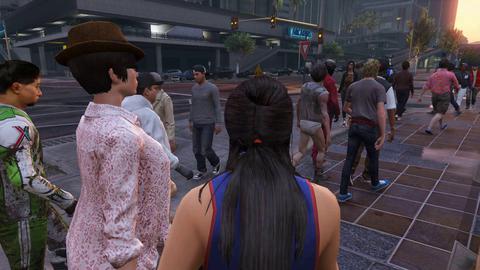}
\end{subfigure}
\begin{subfigure}[t]{0.19\textwidth}
\centering
    \includegraphics[width=\linewidth]{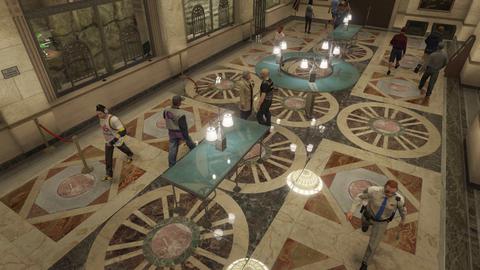}
\end{subfigure}
\caption{Sample images from the MOTSynth dataset \cite{fabbri2021motsynth}. The sequences vary in camera motion and perspective, environment, and lighting.}
\label{fig:motsynth_samples}
\end{figure}

\subsection{Ground Truth} 
In order to create a strong foundation for the evaluation, we are in need of benchmark datasets with consistent annotation standards and leader boards with a wide range of state-of-the-art trackers.
Therefore, we evaluate MOTCOM on the popular MOT17 \cite{milan2016mot16} and MOT20 \cite{dendorfer2020motchallenge} datasets\footnote{With permission from the MOTChallenge benchmark authors.}.
There are seven sequences in the test split of MOT17 and four sequences in the test split of MOT20, some of which are presented in \figref{motchl_samples}.
Furthermore, leader boards are provided for both benchmarks with results from 212 trackers for MOT17 and 80 trackers for MOT20. 
We use the results from the top-30 ranked trackers\footnote{Leader board results obtained on March 4, 2022.} based on the average HOTA score, so as to limit unstable and fluctuating performances.

In order to strengthen and support the evaluation, we include the training split of the fully synthetic MOTSynth dataset \cite{fabbri2021motsynth} which contains 764 varied sequences of pedestrians.
A few samples from the dataset can be seen in \figref{motsynth_samples}.
In order to obtain ground truth tracker performance for MOTSynth, we train and test a CenterTrack model \cite{zhou2020tracking} on the data.
We have chosen CenterTrack as it has been shown to perform well when trained on synthetic data \cite{fabbri2021motsynth}.

\subsection{Evaluation Metrics}
We evaluate and compare the dataset complexity metrics by their ability to rank the MOT sequences according to the HOTA score of the trackers.
We rank the sequences from simple to complex by their \textit{density}, \textit{number of tracks} (abbr. \textit{tracks}), MOTCOM score, and HOTA score.
Depending on the metric, the ranking is in decreasing (HOTA) or increasing order (\textit{density}, \textit{tracks}, MOTCOM).
The absolute difference between the ranks, known as Spearman's Footrule Distance (FD) \cite{diaconis1977spearman}, gives the distance between the ground truth and estimated ranks 
\begin{equation}\label{eq:md}
    \mathrm{FD} = \sum^{n}_{i=1} |\mathrm{rank}(x_i) - \mathrm{rank}(\mathrm{HOTA}_i)|,
\end{equation}
where $n$ is the number of sequences and $x$ is \textit{density}, \textit{tracks}, or MOTCOM.
In order to directly compare results of sets of different lengths, we normalize the FD by the maximal possible distance $\mathrm{FD}_{\mathrm{max}}$ which is computed as 
\begin{equation}
    \mathrm{FD}_{\mathrm{max}} = 
    \begin{cases}
        \sum^{n}_{i=1}i - \frac{n}{2} & \quad \{n ~ | ~ 2m ~, ~ m \in ~ \mathbb{Z}^{+} \}\\
        \sum^{n}_{i=1}i - \frac{n+1}{2} & \quad \{n ~ | ~ 2m-1 ~,~ m\in ~ \mathbb{Z}^{+} \}        
    \end{cases}.
\end{equation}
Finally, we compute the normalized FD, $\mathrm{NFD} = \frac{\mathrm{FD}}{\mathrm{FD}_{\mathrm{max}}}$.

\section{Results}\label{sec:results}
In \tableref{fd_top30}, we present the mean FD of the ranks of \textit{density}, \textit{tracks}, and MOTCOM against the ground truth ranks dictated by the average top-30 HOTA performance on the MOT17 and MOT20 test splits (individually and in combination).
The numbers in parentheses are the normalized FD.
Generally, MOTCOM has a considerably lower FD compared to \textit{density} and \textit{tracks}.
This suggests that MOTCOM is better at ranking the sequences according to the HOTA performance

\begin{table}[b]
\centering  
\caption{Ground truth ranks are based on the average top-30 HOTA performance. The results are presented as the mean FD and the NFD in parentheses. A lower score is better and the results in bold are the lowest}
\label{tab:fd_top30}
\begin{tabular}{@{}lm{1cm}l m{1cm}l m{1cm}l@{}}
\toprule
\textbf{Top-30}  & & MOT17$_\mathrm{test}$         
                 & & MOT20$_\mathrm{test}$         
                 & & Combined \\ \midrule
Density          & & 1.71 (0.50) && 1.00 (0.50) && 3.82 (0.70)  \\
Tracks           & & 2.57 (0.75) && 1.50 (0.75) && 3.82 (0.70)  \\
MOTCOM           & & \textbf{0.86 (0.25)} 
                 & & \textbf{0.00 (0.00)} 
                 & & \textbf{1.45 (0.27)}\\ \bottomrule
\end{tabular}
\end{table}

\begin{table}[t]
\centering
\caption{Ground truth ranks are based on the CenterTrack HOTA performance. The results are presented as the mean FD and the NFD in parentheses. A lower score is better and the results in bold are the lowest}
\label{tab:fd_ct}
\begin{tabular}{@{}lm{0.4cm}l m{0.4cm}l m{0.4cm}l m{0.4cm}l@{}}
\toprule
\textbf{CenterTrack}& & MOTChl$_\mathrm{test}$         
                    & & MOTChl$_\mathrm{train}$         
                    & & MOTChl$_\mathrm{both}$
                    & & MOTSynth \\ \midrule
Density             & & 3.27 (0.60) 
                    & & 4.18 (0.77)
                    & & 7.36 (0.67) 
                    & & 238.71 (0.63)\\
Tracks              & & 2.73 (0.50) 
                    & & 3.64 (0.67) 
                    & & 6.64 (0.60) 
                    & & 193.50 (0.51)\\
MOTCOM              & & \textbf{2.36 (0.43)} 
                    & & \textbf{2.18 (0.40)}
                    & & \textbf{4.82 (0.44)}
                    & & \textbf{100.17 (0.26)}\\ \bottomrule
\end{tabular}
\end{table}

A similar tendency can be seen for the CenterTrack-based results presented in \tableref{fd_ct}.
In order to increase the number of samples, we have evaluated CenterTrack on both the train and test splits of the MOT17 and MOT20 datasets.
MOTChl$_\mathrm{test}$ and MOTChl$_\mathrm{train}$ are the test and train sequences, respectively, of MOT17 \textit{and} MOT20.
MOTChl$_\mathrm{both}$ includes \textit{all} the sequences from MOT17 and MOT20.
These results support our claim that MOTCOM is better at estimating the complexity of MOT sequences compared to \textit{density} and \textit{tracks}.

\begin{wrapfigure}{r}{0.5\textwidth}
    \centering
    \includegraphics[width=\linewidth]{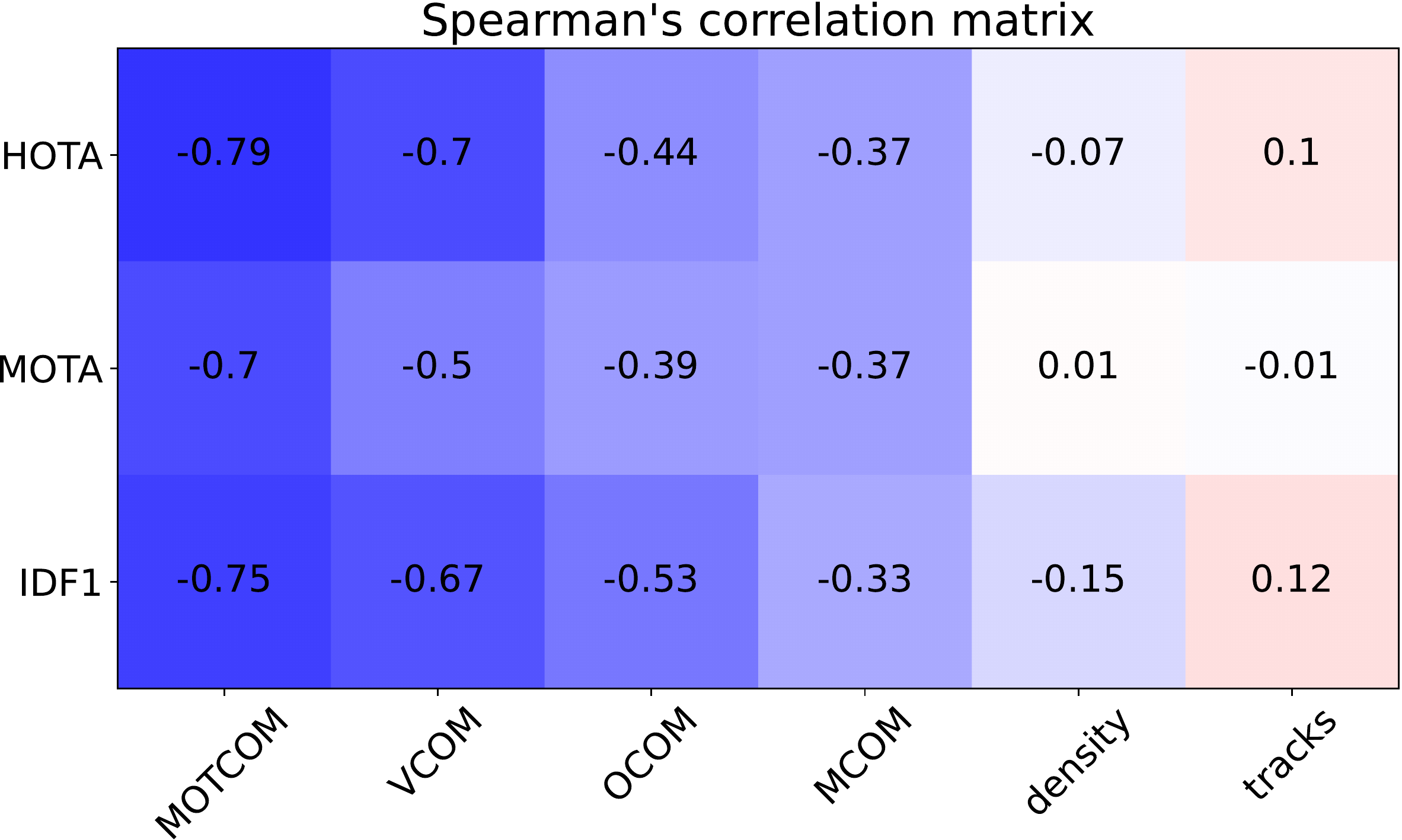}
    \caption{Spearman's correlation matrix based on the performance of the top-30 trackers on MOT17 and MOT20.}
    \label{fig:spearman_corr}
\end{wrapfigure}

We present a Spearman's correlation matrix in \figref{spearman_corr} based on the top-30 trackers evaluated for the combined MOT17 and MOT20 test splits.
It indicates that the \textit{density} and \textit{tracks} do not correlate with HOTA, MOTA, or IDF1, whereas MOTCOM has a strong negative correlation with all the performance metrics.
Trackers evaluated on sequences with high MOTCOM scores tend to have lower performance while sequences with low MOTCOM scores gives higher performance.
This underlines that MOTCOM can indeed be used to understand the complexity of MOT sequences.

\section{Discussion}\label{sec:discussion}
Our complexity metric MOTCOM provides tracker researchers and dataset developers a comprehensive score to investigate and describe the complexity of MOT sequences without the need for multiple baseline evaluations of different tracking methods.
This allows for an objective comparison of different datasets without introducing potential training bias.
Currently, the assessment of tracker performance is roughly speaking reduced to a placement on a benchmark leader board.
This underrates novel solutions developed for less popular datasets or methods designed explicitly to solve sub-tasks such as occlusion or erratic motion.

\begin{figure}[t]
    \centering
    \includegraphics[width=\linewidth]{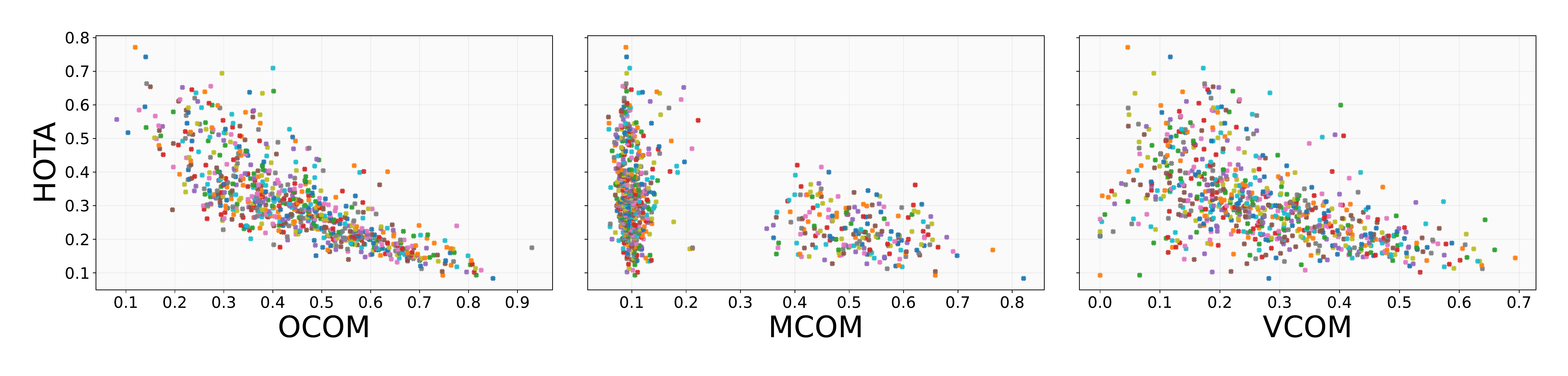}
    \caption{The CenterTrack-based HOTA scores of the MOTSynth sequences plotted against the sub-metrics OCOM, MCOM, and VCOM, respectively.}
    \label{fig:motsynth_sub-metrics}
\end{figure}

\begin{figure}[b]
    \centering
    \includegraphics[width=\linewidth]{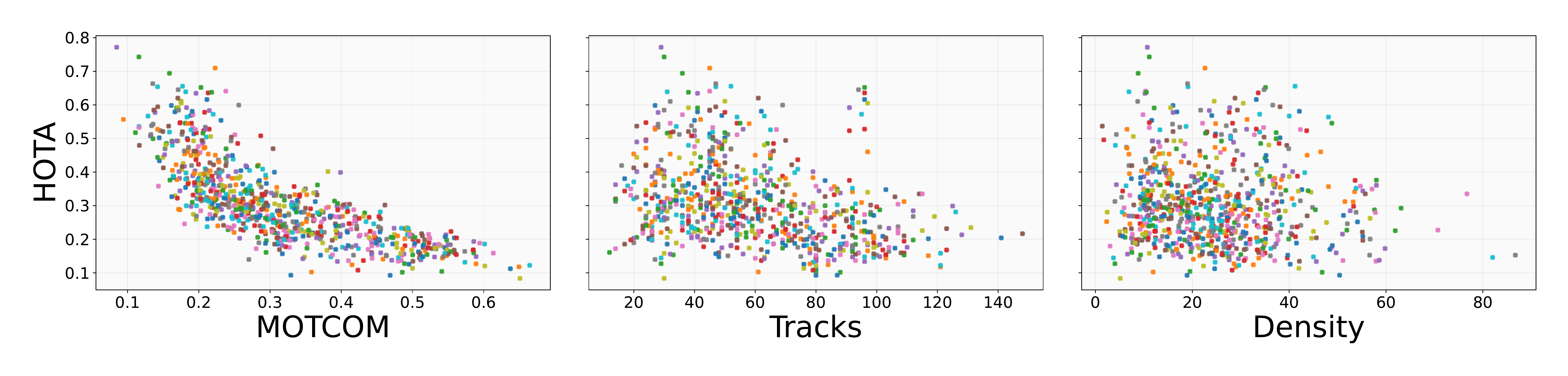}
    \caption{The CenterTrack-based HOTA scores of the MOTSynth sequences plotted against MOTCOM, \textit{tracks}, and \textit{density}.}
    \label{fig:motsynth_results}
\end{figure}

Supplemented by the sub-metrics, MOTCOM provides a deeper understanding and more informed discussions on dataset composition and tracker performance, which will increase the explanability of MOT.
In order to illustrate this, we discuss the performance of CenterTrack on the MOTSynth dataset with respect to MOTCOM.
Here we see that the occlusion level (OCOM) in \figref{motsynth_sub-metrics} has a strong negative correlation with the HOTA score and the visual similarity metric (VCOM) has a relatively weak correlation with HOTA.
Both cases expose the design of CenterTrack, which does not contain a module to handle lost tracks and is not dependent on visual cues for tracking.
For the motion metric (MCOM) we see two distributions; one in the lower end and one in the upper end of the MCOM range.
The objects are expected to behave similarly, so this indicates that parts of the MOTSynth sequences include heavy camera motion which is difficult for CenterTrack to handle.
In \figref{motsynth_results}, we show that MOTCOM is far better at estimating the complexity level compared to \textit{tracks} and \textit{density}.

\section{Conclusion}
We propose MOTCOM, the first meaningful and descriptive MOT dataset complexity metric, and show that it is preferable for describing the complexity of MOT sequences compared to the conventional methods of \textit{number of tracks} and \textit{density}.
MOTCOM is a combination of three individual sub-metrics that describe the complexity of MOT sequences with respect to key obstacles in MOT: occlusion, erratic motion, and visual similarity.
The information provided by MOTCOM can assist tracking researchers and dataset developers in acquiring a deeper understanding of MOT sequences and trackers.
We strongly suggest that the community uses MOTCOM as the prevalent complexity measure for increasing the explainability of MOT trackers and datasets. 

\subsubsection*{Acknowledgements}
This work has been funded by the Independent Research Fund Denmark under case number 9131-00128B.

\clearpage
%
%
\bibliographystyle{splncs04}
\bibliography{egbib}

\newpage
\appendix
\chapter*{Supplementary Materials}

\subsubsection{Computing MOTCOM.}
We have evaluated four averaging methods for combining the sub-metrics into MOTCOM.
The four methods are the arithmetic, quadratic, geometric, and harmonic means and they are presented in \equationref{arithmetic}, \equationref{quadratic},\equationref{geometric}, and \equationref{harmonic}, respectively.
\begin{table*}[!ht]
\begin{tabular}{cc}
\begin{minipage}{0.45\linewidth}
    \begin{equation}\label{eq:arithmetic}
        \mathrm{arithmetic} = \frac{1}{n} \sum^{n}_{i=1} m_i
    \end{equation}
\end{minipage}   
&  
\begin{minipage}{0.45\linewidth}
    \begin{equation}\label{eq:quadratic}
        \mathrm{quadratic} = \sqrt{\frac{1}{n} \sum^{n}_{i=1} m_i^2}
    \end{equation}
\end{minipage}
\\
\begin{minipage}{0.45\linewidth}
    \begin{equation}\label{eq:geometric}
        \mathrm{geometric} = \sqrt[n]{\prod^{n}_{i=1} m_i}
    \end{equation}    
\end{minipage}
    & 
\begin{minipage}{0.45\linewidth}
    \begin{equation}\label{eq:harmonic}
        \mathrm{harmonic} = \frac{n}{\sum^{n}_{i=1} \frac{1}{m_i}}
    \end{equation}   
\end{minipage}
\end{tabular}
\end{table*}

\begin{figure}[]
\centering
    \includegraphics[width=0.6\linewidth]{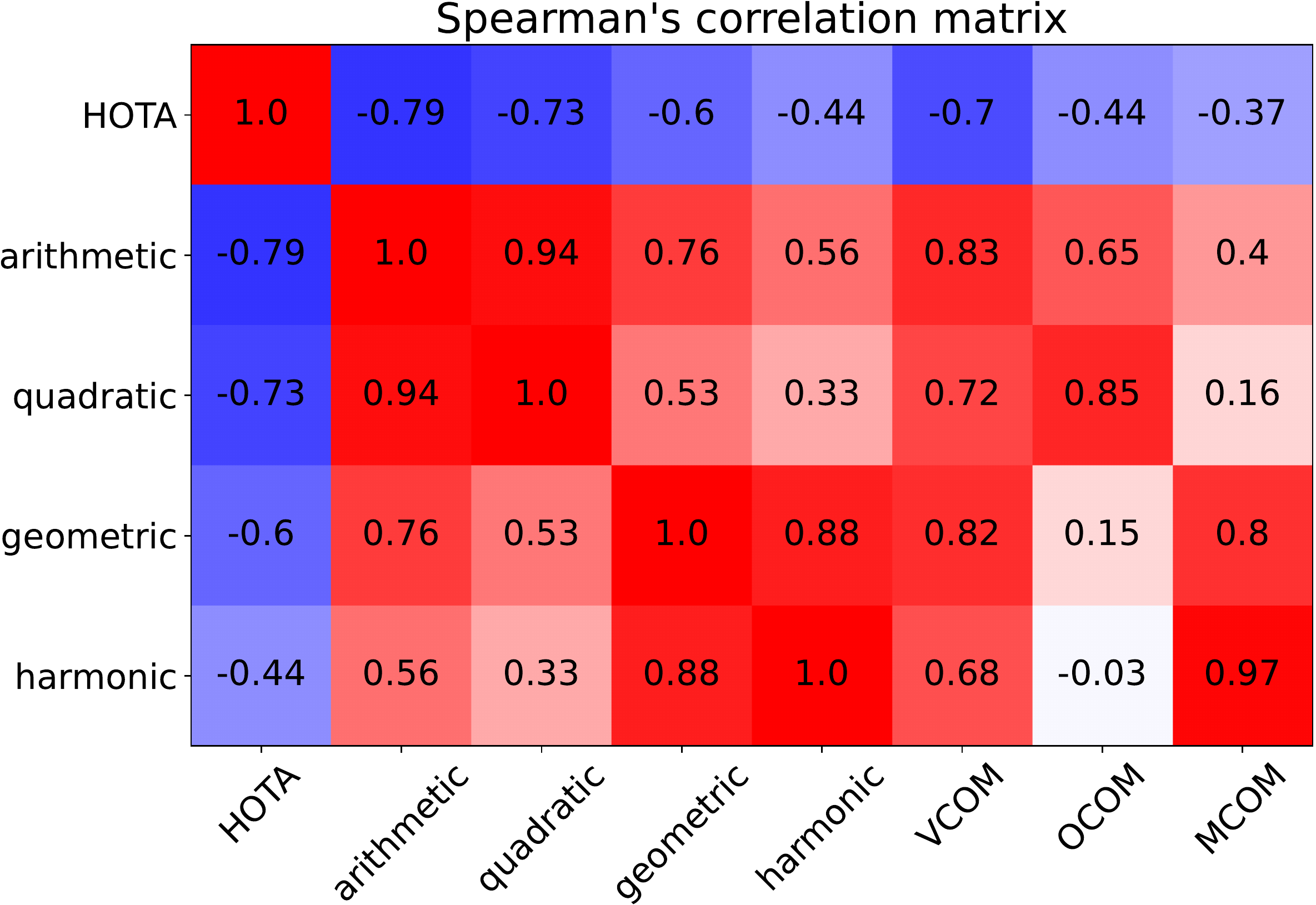}
    \caption{Spearman's correlation matrix. The entries represent the MOTCOM values when the sub-metrics are combined using the four different averaging methods. The HOTA performance is the average of the top-30 ranked trackers. The scores are based on the combined MOT17 and MOT20 test splits.}
    \label{fig:correlation_means}
\end{figure}

We present the four variations of MOTCOM in \figref{correlation_means} computed on the combined MOT17 and MOT20 test splits. 
We see that they all correlate negatively with the HOTA score.
However, the arithmetic mean has the strongest negative correlation and it correlates positively with all the sub-metrics.
Therefore, we suggest to compute MOTCOM as the arithmetic mean of the sub-metrics.\clearpage
\subsubsection{Complexity Score Plots for MOT17 and MOT20}
\begin{figure}[]
    \centering
    \begin{subfigure}[b]{0.55\textwidth}
    \centering
    \includegraphics[width=\linewidth]{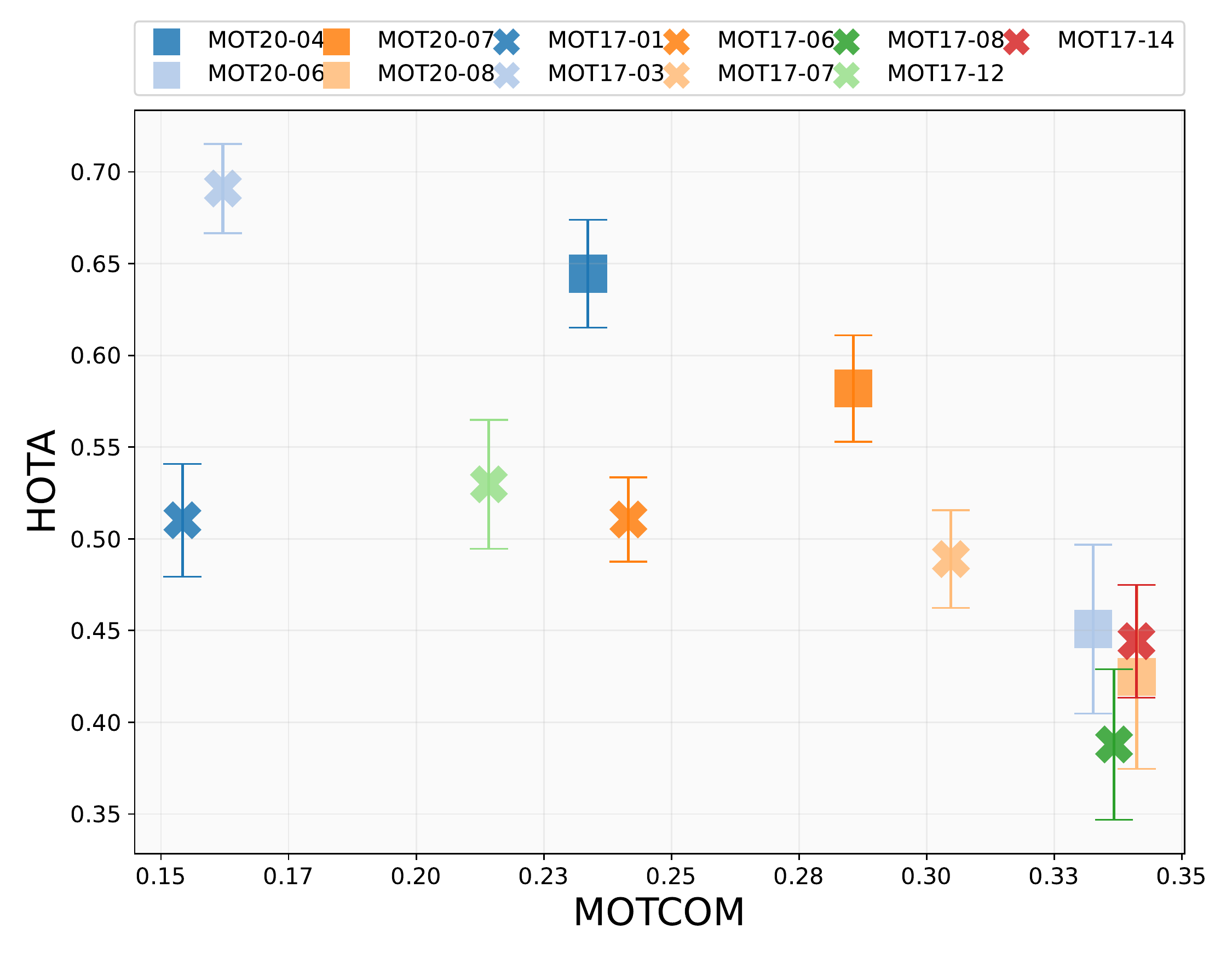}
    \caption{MOTCOM vs. HOTA.}
    \label{fig:motcom_hota_top30}
    \end{subfigure}
    ~
    \begin{subfigure}[b]{0.45\textwidth}
    \centering
    \includegraphics[width=\linewidth]{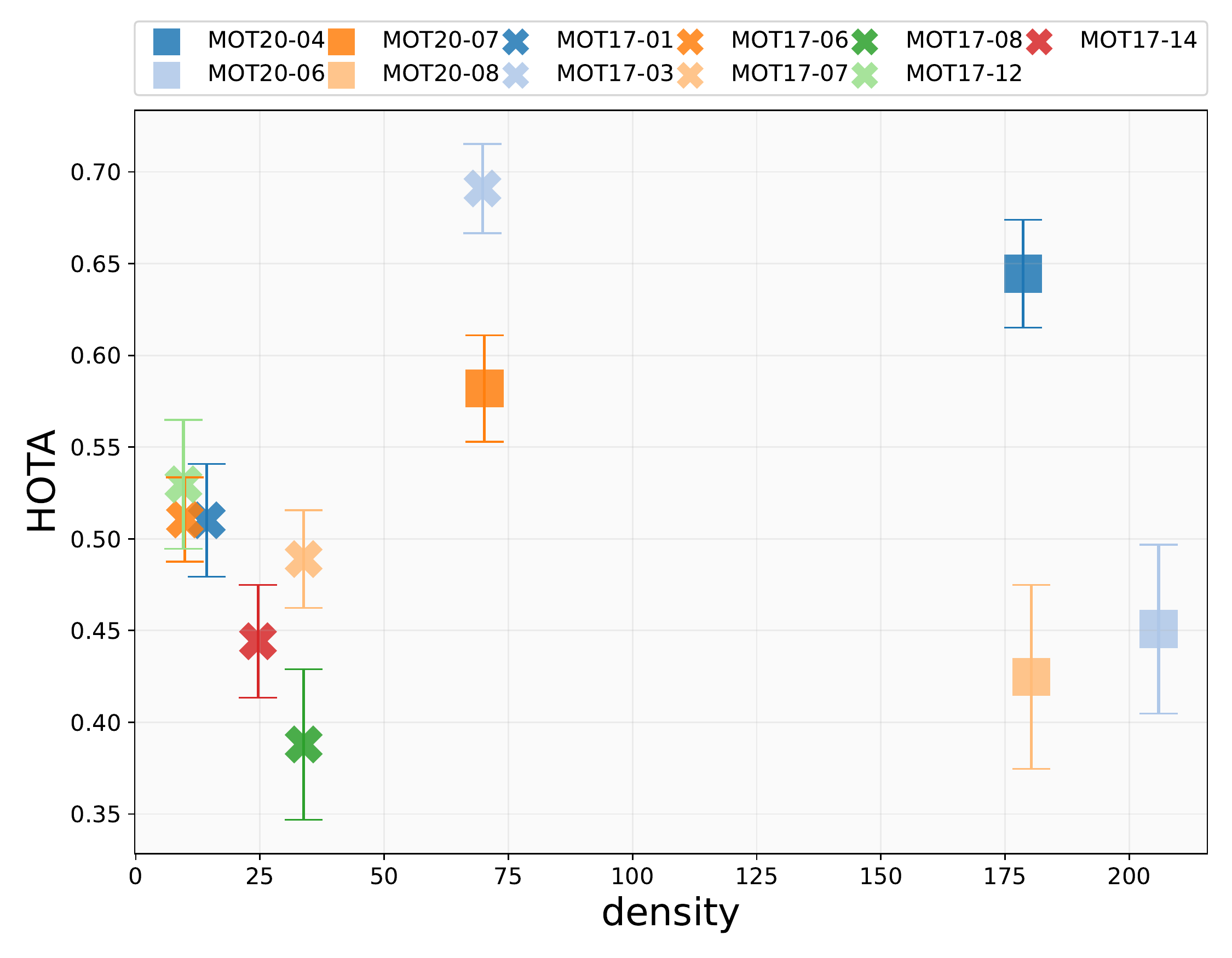}
    \caption{Density vs. HOTA.}
    \label{fig:density_hota_top30}
    \end{subfigure}
    ~
    \begin{subfigure}[b]{0.45\textwidth}
    \centering
    \includegraphics[width=\linewidth]{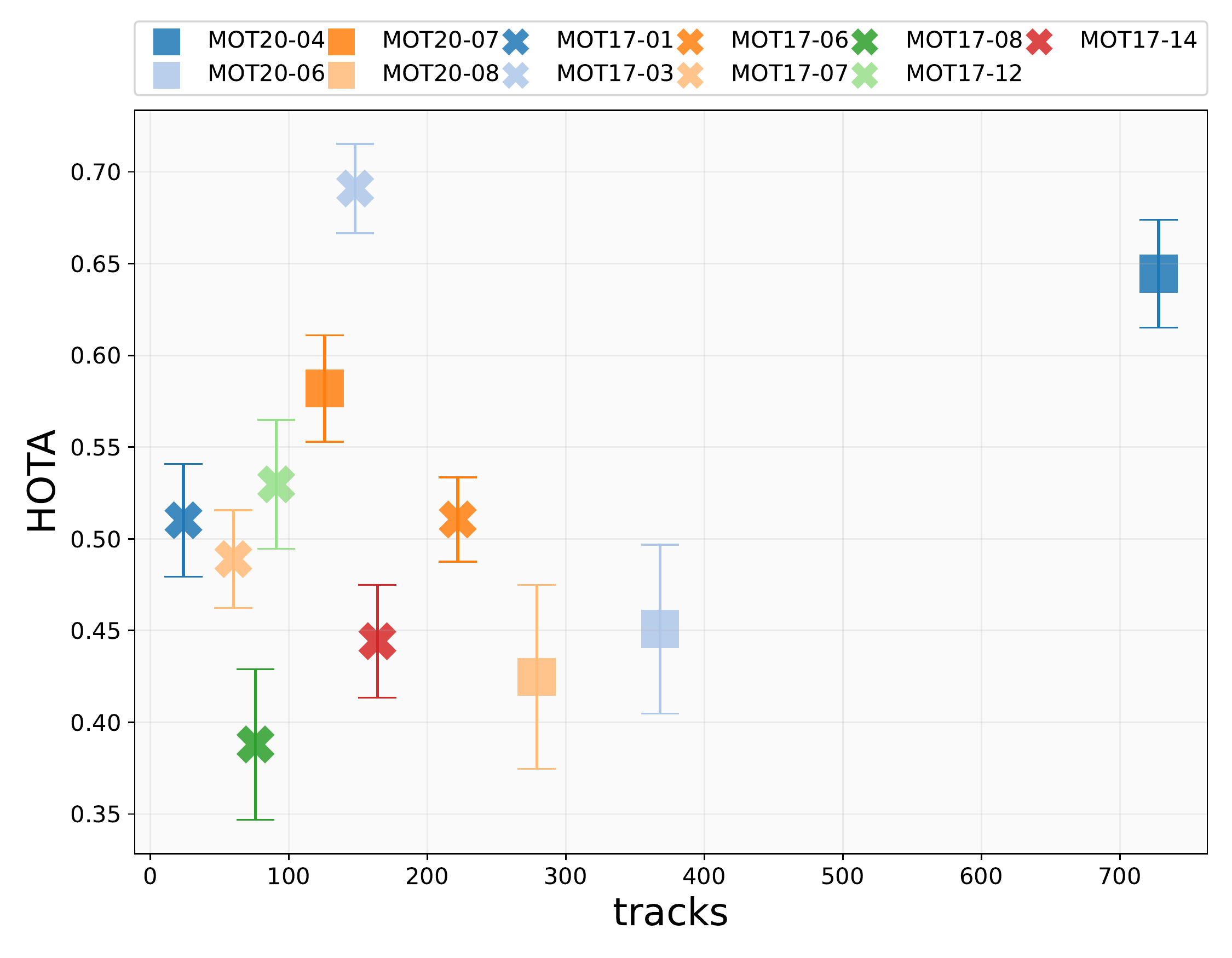}
    \caption{Tracks vs. HOTA.}
    \label{fig:tracks_hota_top30}
    \end{subfigure}
    \caption{Average HOTA performance of the top-30 trackers on MOT17 and MOT20 test split against a) MOTCOM, b) \textit{density}, and c) \textit{tracks}. Square markers represent MOT20 sequences and crosses are MOT17 sequences.}
    \label{fig:complexity_metrics_hota}
\end{figure}

In the main paper we evaluate MOTCOM, \textit{density}, and \textit{tracks} on the MOT17 and MOT20 test splits.
We focus mainly on the ranking capabilities of the metrics as we expect tracker performance to have a monotonic, but not necessarily linear, relationship with complexity.
The ranks of MOTCOM, \textit{density}, and \textit{tracks} presented in the main paper are based on the scores displayed in \figref{complexity_metrics_hota}.

The position of the marker indicates the average score and the error bar is the standard deviation.
The marker of the MOT17 sequences is a cross and the MOT20 sequences are represented by a square.
We see that the MOT20 sequences have significantly higher densities and more tracks compared to the MOT17 sequences, but the HOTA performance is not correspondingly low.
This illustrates that \textit{density} and \textit{tracks} do not suffice to describe the complexity of MOT sequences.
\subsubsection{Complete Spearman's Correlation Matrix for MOT17 and MOT20.}
\begin{figure}[]
\centering
    \centering
    \includegraphics[width=0.75\linewidth]{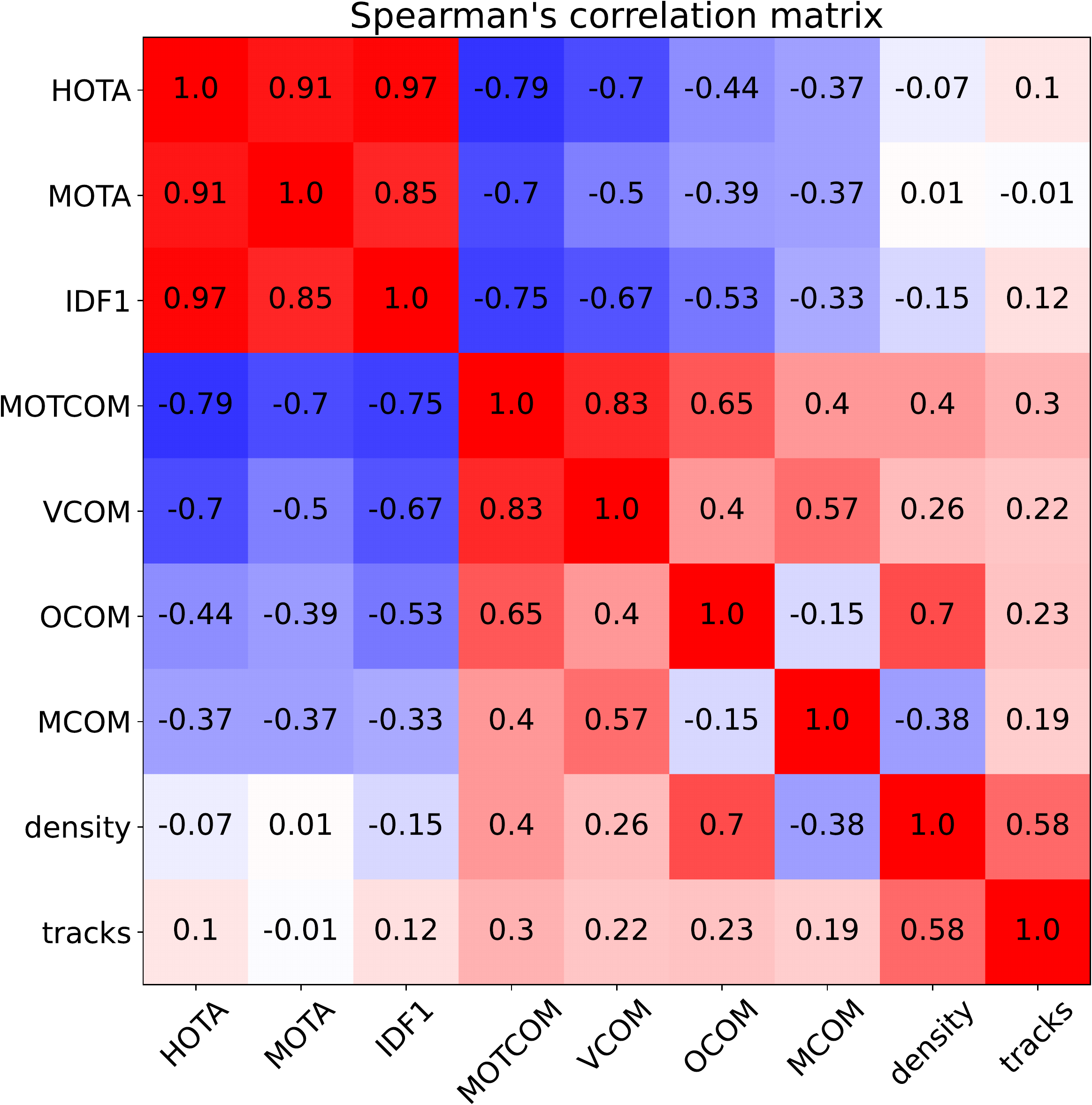}
    \caption{Spearman's correlation matrix. Based on the average performance of the top-30 trackers on MOT17 and MOT20 test split.}
    \label{fig:spearman_top-30}
\end{figure}

In Figure 9 in the main paper we presented a partial Spearman's correlation matrix based on the MOT17 and MOT20 sequences.
We used the matrix to evaluate the monotonic relationship between the three complexity metrics (MOTCOM, \textit{density}, and \textit{tracks}) and HOTA, MOTA, and IDF1.
In \figref{spearman_top-30} we present the complete Spearman's correlation matrix, which shows additional details on the relationship between the entries.

\newpage

\subsubsection{Complete Spearman's Correlation Matrix for MOTSynth.}
\begin{figure}[]
        \centering
    \includegraphics[width=0.75\linewidth]{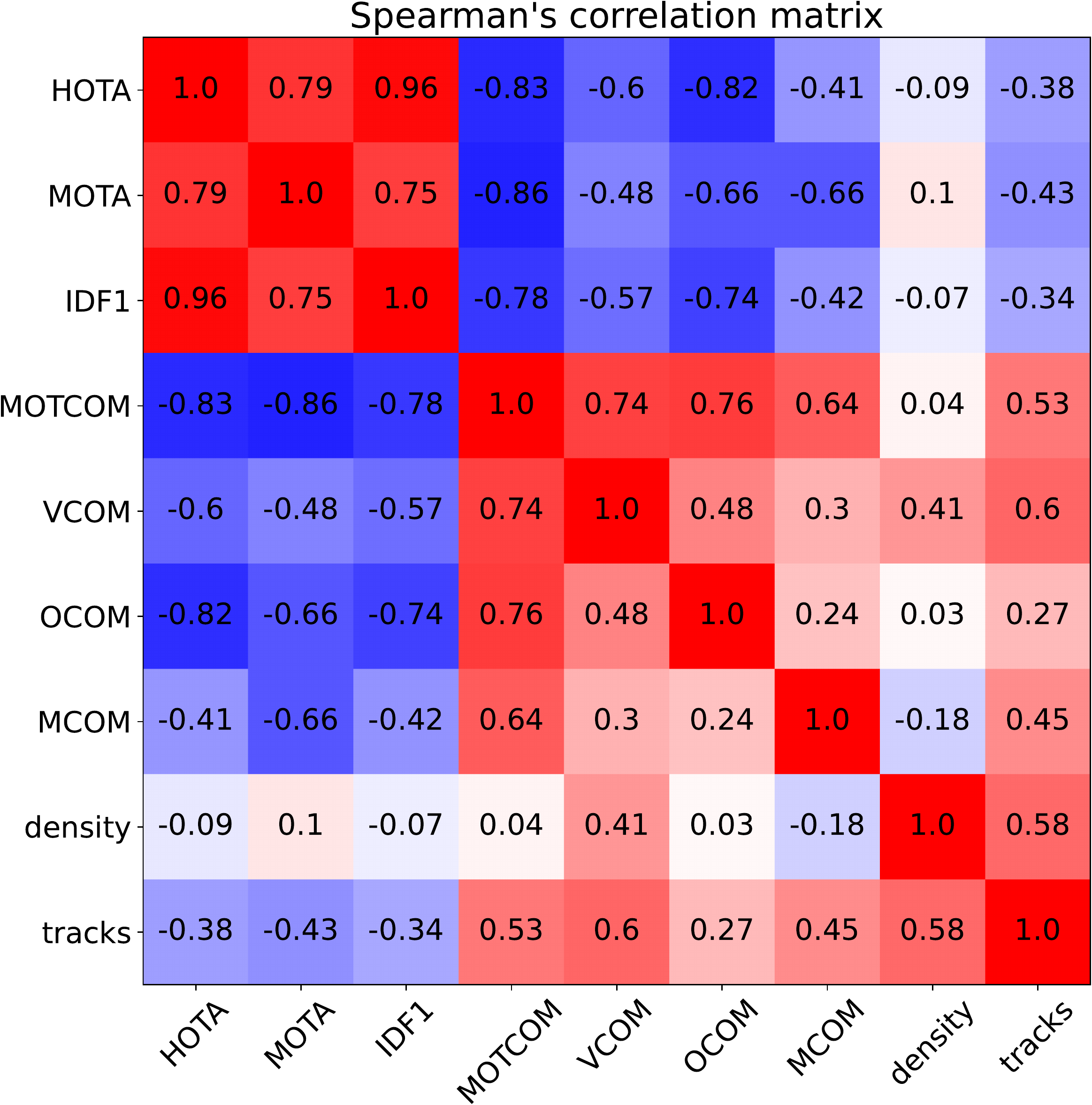}
    \caption{Spearman's correlation matrix. Based on the CenterTrack performance on the MOTSynth train split.}
    \label{fig:spearman_motsynth}
\end{figure}

In the main paper we presented the Spearman's Footrule Distance and the complexity scores for the MOTSynth sequences.
To expand upon this, we include the complete Spearman's correlation matrix for the MOTSynth train split in \figref{spearman_motsynth}.
The matrix gives a detailed overview of the monotonic relationship between the entries.

\end{document}